\documentclass[conference]{IEEEtran}
\IEEEoverridecommandlockouts
\usepackage[noadjust]{cite}
\usepackage{amsmath,amssymb,amsfonts}
\usepackage{algorithmic}
\usepackage{graphicx}
\usepackage{textcomp}
\usepackage{xcolor}
\usepackage{times}
\usepackage{multirow}
\usepackage{multicol}
\usepackage{tabularx}
\usepackage{pgfplots}
\usepackage{array}
\def\BibTeX{{\rm B\kern-.05em{\sc i\kern-.025em b}\kern-.08em
    T\kern-.1667em\lower.7ex\hbox{E}\kern-.125emX}}

\makeatletter
\renewcommand\IEEEkeywordsname{Keywords}
\makeatother
\usepackage{xpatch}
\xpatchcmd\IEEEkeywords{---}{-}{}{}

\makeatletter
\renewcommand{\fnum@figure}{Figure~\thefigure}
\makeatother

\begin{document}

\title{\bfseries\Large Sarcasm Detection as a Catalyst: Improving Stance Detection with Cross-Target Capabilities

}

\author{\IEEEauthorblockN{Gibson Nkhata, Shi Yin Hong, Susan Gauch}
\IEEEauthorblockA{\textit{Department of Electrical Engineering \& Computer Science} \\
\textit{University of Arkansas}\\
Fayetteville, AR 72701, USA \\
Emails: gnkhata@uark.edu, syhong@uark.edu, sgauch@uark.edu}
}

\maketitle

\begin{abstract}
Stance Detection (SD) in social media has become a critical area of interest due to its applications in social, business, and political contexts, leading to increased research within Natural Language Processing (NLP). However, the subtlety, nuance, and complexity of texts sourced from online platforms, often containing sarcasm and figurative language, pose significant challenges for SD algorithms in accurately determining the author's stance. This paper addresses these challenges by employing sarcasm detection as an intermediate-task transfer learning approach specifically designed for SD. Additionally, it tackles the issue of insufficient annotated data for training SD models on new targets by conducting many-to-one Cross-Target SD (CTSD). The proposed methodology involves fine-tuning BERT and RoBERTa models, followed by sequential concatenation with convolutional layers, Bidirectional Long Short Term Memory (BiLSTM), and dense layers. Rigorous experiments are conducted on publicly available benchmark datasets to evaluate the effectiveness of our transfer-learning framework. The approach is assessed against various State-Of-The-Art (SOTA) baselines for SD, demonstrating superior performance. Notably, our model outperforms the best SOTA models in both in-domain SD and CTSD tasks, even before the incorporation of sarcasm-detection pre-training. The integration of sarcasm knowledge into the model significantly reduces misclassifications of sarcastic text elements in SD, allowing our model to accurately predict 85\% of texts that were previously misclassified without sarcasm-detection pre-training on in-domain SD. This enhancement contributes to an increase in the model's average macro F1-score. The CTSD task achieves performance comparable to that of the in-domain task, despite using a zero-shot fine-tuning approach, curtailing the lack of annotated samples for training unseen targets problem. 
Furthermore, our experiments reveal that the success of the transfer-learning framework depends on the correlation between the lexical attributes of the intermediate task (sarcasm detection) and the target task (SD). This study represents the first exploration of sarcasm detection as an intermediate transfer-learning task within the context of SD, while also leveraging the concatenation of BERT or RoBERTa with other deep-learning techniques. The proposed approach establishes a foundational baseline for future research in this domain.
\end{abstract}

\begin{IEEEkeywords}
\textbf{\textit{Stance detection; sarcasm detection; transfer learning; BERT; RoBERTa.}}
\end{IEEEkeywords}

\section{Introduction}

This paper extends our previous research on intermediate-task transfer learning, specifically, leveraging sarcasm detection to enhance Stance Detection (SD)~\cite{gnkhata1}. In our prior work, we focused on pretraining models on sarcasm detection before fine-tuning them on SD, utilizing in-domain training data of SD targets.  This study further explores SD from two perspectives: in-domain SD, where a single target is used for training and evaluation, and Cross-Target SD (CTSD), which involves training a model on one or more targets and evaluating it on a different target. CTSD represents the latest research direction in this area.

The proliferation of the Internet and social media platforms such as Twitter (X), Facebook, microblogs, discussion forums, and online reviews has significantly altered how individuals communicate and share information~\cite{savini2022intermediate}\cite{grvcar2017stance}. These platforms allow users to express opinions and engage with global audiences on various topics, including current trends, products, and politics~\cite{newman2011mainstream,aldayel2021stance,ghosh2019stance}. The vast amount of discourse generated on these platforms provides valuable data for Natural Language Processing (NLP) tasks, particularly SD.

 SD is the automated identification of an individual's stance based solely on their utterance or written material~\cite{ghosh2019stance}\cite{augenstein2016stance,kuccuk2020stance,kuccuk2022tutorial}. Stance refers to the expression of a speaker's or author's position, attitude, or judgment toward a specific topic, target, or proposition~\cite{aldayel2021stance}\cite{biber1988adverbial}. Stance labels typically categorize expressions into \textit{InFavor}, \textit{Against}, or \textit{None}. SD has become increasingly relevant in various domains such as opinion mining, fake news detection, rumor verification, election prediction, information retrieval, and text summarization~\cite{aldayel2021stance}\cite{biber1988adverbial}.

SD research can be broadly classified into two perspectives~\cite{aldayel2021stance}: detecting expressed views and predicting unexpressed views. The former involves categorizing an author's text to determine their current stance toward a given subject~\cite{aldayel2021stance}\cite{mohammad2016semeval}, while the latter aims to infer an author's position on an event or subject that they have not explicitly discussed~\cite{magdy2016isisisnotislam}\cite{darwish2018predicting}. Additionally, SD tasks can be categorized as either Target-Specific SD (TSSD) or Multi-Target SD (MTSD). TSSD focuses on individual subjects, whereas MTSD involves jointly inferring stances toward multiple related subjects~\cite{ghosh2019stance}\cite{aldayel2021stance}\cite{wei2018multi}\cite{sobhani2019exploring,sobhani2017dataset,liu2018two}. This paper primarily addresses detecting expressed views within the TSSD framework, incorporating unexpressed views through the infusion of sarcasm knowledge into the model framework. Examples of the SD task are provided in Table \ref{tab:sd_ex}.

\renewcommand{\arraystretch}{1.2} 

\begin{table*}[h]
\label{tab:sd_ex}
\centering
\caption{Examples of the Stance Detection Task}
\label{tab:example}
\begin{tabularx}{0.75\textwidth}{|p{2.5cm}|X|p{2cm}|}
\hline
\textbf{Target} &\textbf{Text} & \textbf{Stance} \\
\hline
Feminist Movement & Women don't make 75\% less than men for the same job. Women, on average, make less than men. Look it up feminazis. \#EqualPayDay \#SemST
 &Against \\
\hline
Feminist Movement &Congratulations to America for overcoming 1 battle for \#equality. Now let's have women \& all races treated equally \#AllLivesMatter \#SemST
 & Favor \\
\hline
Feminist Movement &Honoured to be followed by the truly inspirational \@Kon\_\_K founder of @ASRC1 \#realaustralianssaywelcome \#thethingsthatmatter \#SemST
 & None \\
\hline
\end{tabularx}
\end{table*}

Previous SD research has primarily utilized publicly available datasets sourced from online platforms~\cite{ghosh2019stance}\cite{aldayel2021stance}\cite{kuccuk2020stance}\cite{kuccuk2021stance}. However, texts from these platforms often exhibit subtlety, nuance, and complexity, including sarcastic and figurative language. These characteristics present challenges for SD algorithms in accurately discerning the author's stance~\cite{ghosh2019stance}. Moreover, targets are not always explicitly mentioned in the text~\cite{augenstein2016stance}, and stances may not be overtly expressed, further complicating the task of inferring the author's stance. 
Due to this problem, some examples discussed do not necessarily reflect the authors' beliefs. This often requires implicit inference through a combination of interactions, historical context, and sociolinguistic attributes such as sarcasm or irony.

To address these challenges, prior work has explored intermediate-task transfer learning, involving the fine-tuning of a model on a secondary task before its application to the primary task~\cite{savini2022intermediate}\cite{phang2018sentence,li2019multi,sap2019socialiqa,pruksachatkun2020intermediate,hardalov2022few}. For instance,~\cite{li2019multi} and~\cite{hardalov2022few} utilized sentiment classification to enhance their models for SD. Similarly,~\cite{savini2022intermediate} incorporated emotion and sentiment classification prior to sarcasm detection, suggesting that pre-training with sentiment analysis before sarcasm detection improves overall performance due to the correlation between sarcasm and negative sentiment. This finding aligns with one of our experimental observations in Section \ref{Exp}, where most sarcastic sentences with an ``Against" stance are initially misclassified as ``InFavor" before incorporating sarcasm pre-training into our model. However, despite its potential, sarcasm has been relatively unexplored as a means of improving SD models. In this study, we experiment with sarcasm detection as an intermediate task tailored to enhance SD performance.

Sarcasm detection involves discounting literal meaning to infer intention or secondary meaning from an utterance~\cite{ghosh2016fracking}. Sarcasm often involves using positive words or emotions to convey negative, ironic, or figurative meanings~\cite{sarsam2020sarcasm}\cite{jamil2021detecting}. For example, in the sarcastic sentence ``\textit{I like girls. They just need to know their place}," the word ``like" is used figuratively to mock the subject, making it difficult for SD algorithms to detect the true stance without accounting for sarcasm. Thus, sarcasm can alter the stance of a text from \textit{Against} to \textit{InFavor} and vice versa if not properly addressed~\cite{jamil2021detecting}\cite{liebrecht2013perfect}. Based on these observations, we developed an SD approach that incorporates sarcasm detection.

This study employs a model framework consisting of BERT~\cite{devlin2018bert} or RoBERTa~\cite{liu2019roberta}, convolutional layers (Conv), a Bidirectional Long Short Term Memory (BiLSTM) layer, and a dense layer. Our experimental results demonstrate the efficacy of this approach, evidenced by improved macro F1-scores when sarcasm detection is included in the model framework. Additionally, we explore the impact of different sarcasm detection approaches on SD performance, considering the linguistic and quantitative attributes inherent in sarcasm datasets. Furthermore, the significance of this approach is underscored through a failure analysis of sarcastic texts from datasets, revealing the limitations of the original SD model before sarcasm pre-training.

We extend the work from~\cite{gnkhata1} by applying CTSD to our tasks using a leave-one-out training approach. This method explores zero-shot fine-tuning on the target of interest, where four targets are used for model training and the remaining one for evaluation. The goal is to transfer knowledge from other targets to the target with limited training examples, thereby circumventing the scarcity of training data and the challenges of annotating sufficient data for new targets~\cite{zhang-etal-2020-enhancing-cross}.

CTSD can be approached in two traditional ways: one-to-one, where one source target is used for training and one destination target for evaluation, and many-to-one, where multiple source targets are used for training and one destination target for evaluation~\cite{zhang-etal-2020-enhancing-cross,xu-etal-2018-cross,khiabani2023fewshotlearningcrosstargetstance,9746302}. The former approach often underutilizes available targets and struggles with generalization to unrelated targets, while the latter addresses these issues but often relies on sophisticated meta-learning approaches and limited datasets have been explored. In this work, we explore the many-to-one CTSD approach on two competitive SD tasks, proposing a solution that integrates sarcasm detection while mitigating the challenges associated with limited annotated SD data through many-to-one CTSD on diverse datasets.

Our experimental results show that the cross-target approach achieves performance comparable to models trained on target-specific data. Further analysis, including correlation measures between training and evaluation targets using cosine similarity on pre-trained language model embeddings, suggests that the overlapping vocabulary between the targets contributes to this performance.

Our work makes the following key contributions:

\begin{itemize}
\item
\textit{Transfer-Learning Framework:} Introducing a novel transfer-learning framework incorporating sarcasm detection as an intermediate task before fine-tuning on SD, utilizing an integrated deep learning model.

\item 
\textit{Cross-Target Stance Detection:} Introducing the leave-one-out fine-tuning on the SD targets, using four targets in training and the remaining one during evaluation, giving performance on par with training on the latter's designated data and curtailing the lack of annotated samples for training unseen targets problem. 

\item
\textit{Performance Superiority:} Demonstrating superior performance against State-Of-The-Art (SOTA) SD baselines, even without sarcasm detection pre-training, as indicated by higher macro F1-scores.

\item
\textit{Correlation Analysis:} Establishing and illustrating the correlation between sarcasm detection and SD, exemplified through a failure analysis, thereby emphasizing the improvement of SD through sarcasm detection.

\item
\textit{Impact Assessment:} Measuring the impact of various sarcasm detection models on target tasks based on the correlation between linguistic and quantitative attributes in the datasets of the two tasks.

\item
\textit{Ablation Study:} Conducting an ablation study to assess the contribution of each module to the overall model framework. The study also reveals a significant drop in performance without sarcasm knowledge, underscoring the importance of our proposed approach.
\end{itemize}

The remainder of this paper unfolds as follows: Section~\ref{Rel} reviews related work, Section~\ref{Meth} outlines our proposed approach, and Section~\ref{Exp} delves into comprehensive experiments, covering datasets, results, and subsequent discussions.  The limitations inherent in our study are critically examined in Section~\ref{limit}. The final section provides the conclusion and recommendations for further research.

\section{Related Work} \label{Rel}
This section comprehensively reviews the literature on  SD and intermediate-task transfer learning.

\subsection{Stance Detection (SD)}
The research on SD has traditionally been explored from two primary perspectives: Target-Specific SD (TSSD), which focuses on individual targets~\cite{ghosh2019stance}\cite{aldayel2021stance}\cite{zhang2022would}\cite{liang2022zero}, and Multi-Target SD (MTSD), which aims to infer stances toward multiple related subjects concurrently~\cite{liang2022zero}\cite{sobhani2017dataset, liu2018two, sobhani2019exploring}. Early approaches to SD were based on rule-based methods~\cite{zhang2022would}\cite{walker2012stance}, followed by classical machine learning techniques~\cite{kuccuk2018stance, segura2018labda, hasan2013stance, mohammad2017stance}. For instance,~\cite{hasan2013stance} applied Naive Bayes (NB) to SD using datasets and features derived from inter-post constraints in online debates. Similarly,~\cite{kuccuk2018stance} utilized features such as unigrams, bigrams, hashtags, external links, emoticons, and named entities in various Support Vector Machine (SVM) models, while~\cite{mohammad2017stance} employed an SVM model with linguistic (n-grams) and sentiment features to predict stance. In contrast,~\cite{segura2018labda} explored and compared linear SVM, Logistic Regression (LR), Multinomial NB, k-Nearest Neighbors (kNN), Decision Trees (DT), and Random Forests (RF) using the simple Bag-of-Words approach with term frequency-inverse document frequency (tf-idf) vectors of tweets as features for multi-modal SD.

While classical approaches relied on manually crafted features, the advent of deep learning models has seen neural networks gradually replace traditional methods~\cite{augenstein2016stance}\cite{ hardalov2022few}\cite{siddiqua2019tweet}\cite{ng2022my}. For instance, the work by~\cite{augenstein2016stance} investigated SD using Bidirectional Conditional Encoding (BCE)~\cite{graves2005framewise}, incorporating an LSTM architecture to build a tweet representation dependent on the target. Similarly,~\cite{wei2016pkudblab} employed a CNN for SD, incorporating a voting scheme mechanism, while~\cite{li2019multi} utilized a bidirectional Gated Recurrent Unit (biGRU) within a multi-task framework that included a target-specific attention mechanism, leveraging sentiment classification to enhance SD performance. Moreover,~\cite{siddiqua2019tweet} presented a neural ensemble model combining BiLSTM, an attention mechanism, and multi-kernel convolution, evaluated on both TSSD and MTSD. Although our work shares some similarities in model framework, it uniquely employs BERT or RoBERTa and introduces an intermediate-task transfer learning technique, diverging from ensemble approaches and multi-kernel usage.

Deep learning models necessitate large datasets for effective SD model training and generalization~\cite{devlin2018bert}. Consequently, recent research has explored the use of pre-trained language models for SD. For example,~\cite{ng2022my} proposed using BERT~\cite{devlin2018bert} in a cross-validation approach, developing a multi-dataset model from the aggregation of several datasets. Similarly,~\cite{ghosh2019stance} conducted a comparative study, fine-tuning pre-trained BERT against classical SD approaches, while~\cite{liang2022zero} employed BERT as an embedding layer to encode textual features in a zero-shot deep learning setting, yielding promising results. On the other hand,~\cite{zhang2022would} experimented with ChatGPT, directly prompting the model with test cases to discern stances; however, all these studies reported difficulties in accurately classifying sarcastic examples.

\subsection{Cross-Target Stance Detection (CTSD)}
Research on CTSD can be divided into two main approaches. The first is the one-to-one approach, where a single source target and a single destination target share common words, which helps bridge the knowledge gap~\cite{zhang-etal-2020-enhancing-cross, xu-etal-2018-cross, khiabani2023fewshotlearningcrosstargetstance}. For example,~\cite{xu-etal-2018-cross} introduced the CrossNet model, utilizing an aspect attention layer to learn domain-specific aspects from a source target for generalization on a destination target. Similarly,~\cite{zhang-etal-2020-enhancing-cross} used external knowledge, such as semantic and emotion lexicons, to enable knowledge transfer between targets. Meanwhile,~\cite{khiabani2023fewshotlearningcrosstargetstance} explored few-shot learning by leveraging social network features alongside textual content, introducing 300+ training examples from the destination target. This line of research primarily explores related targets within a common domain.

The second approach is the many-to-one method, which involves using multiple source targets for a single destination target. For instance,~\cite{9746302} used many unrelated source targets to the destination target without leveraging external knowledge but instead employed a sophisticated meta-learning approach and did not utilize diverse datasets.

\subsection{Intermediate-Task Transfer Learning}
Recent studies have increasingly adopted intermediate-task transfer learning, which transfers knowledge from a data-rich auxiliary task to a primary task~\cite{pruksachatkun2020intermediate}. This technique has proven highly effective across various NLP tasks. For example,~\cite{phang2018sentence} employed supervised pre-training on four-example intermediate tasks to enhance performance on primary tasks evaluated using the GLUE benchmark suite~\cite{wang2018glue}. Additionally,~\cite{hardalov2022few} introduced few-shot learning, leveraging sentiment-based annotation to improve cross-lingual SD performance. Furthermore,~\cite{savini2022intermediate} employed transfer learning by sequentially fine-tuning pre-trained BERT on emotion and sentiment classification before applying it to sarcasm detection, capitalizing on the correlation between sarcasm and negative sentiment polarity.

To the best of our knowledge, prior research has not explored sarcasm detection pre-training for SD, nor has it investigated the concatenation of BERT or RoBERTa with other deep learning techniques for SD. In this paper, we propose leveraging sarcasm detection for both in-domain SD and CTSD within a model framework comprising BERT, convolutional layers, BiLSTM, and a dense layer.

\section{Methodology} \label{Meth}
This section delineates our approach, covering problem formulation, intermediate-task transfer learning, and the model architecture.

\subsection{Problem Formulation} 

We denote the collection of labeled data in the source targets as \(X^{s} = \{x^{s}_{i}, y^{s}_{i}, t^{sj}_{i}\}^{N}_{i=1}, j= \{1,2,3,...,k\}\), where \(x\) represents the input text, \(y\) denotes the stance label, and \(t\) indicates the \(j^{th}\) target. Here, \(s\) represents a source target, and there are \(k\) source targets in \(X^{s}\), comprising \(N\) data samples in total. Similarly, we denote the collection of data in the destination target as \(X^{d} = \{x^{d}_{i}, y^{d}_{i}, t^{d}_{i}\}^{M}_{i=1}, d=\{1\}\), where \(d\) represents the destination target, with \(M\) data samples. Given an input text \(x\) from a destination target \(t^d\), the objective is to predict the stance label of \(x\) towards \(t^{d}\) using the model trained on the labeled data \(X^{s}\). For the in-domain task, \(t^s = t^d\); for the CTSD task, \(t^s \neq t^d\).

\subsection{Intermediate-Task Transfer Learning}

Our approach incorporates intermediate-task transfer learning, which involves two phases: pre-training on an intermediate task and fine-tuning on a target task.

\subsubsection{Target Task}
The primary task in this study is SD, aiming to predict the stance expressed in a given text, such as a tweet, towards a specific target (e.g., `\textit{feminist movement}`). A tweet, denoted as \(x\), is represented as a sequence of words \((w_{1}, w_{2}, w_{3}, \ldots, w_{L})\), with \(L\) representing the sequence length. Stance labels are categorized as \textit{InFavor} (supporting the target), \textit{Against} (opposing the target), or \textit{None} (neutral towards the target).

\subsubsection{Intermediate Task}
The intermediate task in this study is sarcasm detection, where the goal is to determine whether a given text \(S\) is sarcastic. Sarcasm detection labels are categorized as \textit{Sarcastic} (the text is sarcastic) or \textit{Non-Sarcastic} (the text is not sarcastic). As sarcasm has not previously been employed as an intermediate task, we explore three sarcasm-detection datasets to identify key linguistic features that can enhance SD performance:

\textit{Sarcasm V2 Corpus (SaV2C).} The SaV2C dataset, introduced by \cite{oraby2017creating}, is a diverse corpus developed using syntactical cues and crowd-sourced from the Internet Argument Corpus (IAC 2.0). It comprises 4,692 lines containing quote and response sentences from political debates in IAC online forums. SaV2C is categorized into: 1) General Sarcasm (Gen, 3,260 sarcastic and 3,260 non-sarcastic comments); 2) Rhetorical Questions (RQ, 851 rhetorical and 851 non-rhetorical questions); and 3) Hyperbole (Hyp, 582 hyperboles and 582 non-hyperboles). Our focus is on the General Sarcasm category, which includes 3,260 sarcastic and 3,260 non-sarcastic comments.

\textit{The Self-Annotated Reddit Corpus (SARC).} Created by \cite{khodak2018large}, the SARC dataset contains over a million sarcastic and non-sarcastic statements from Reddit. This dataset features a balanced ratio of sarcastic and non-sarcastic comments, with 1,010,826 training and 251,608 evaluation statements. We utilized the Main Balanced variant, obtained directly from the author of \cite{savini2022intermediate}.

\textit{SARCTwitter (ST).} Released by \cite{mishra2016predicting}, the ST dataset includes 350 sarcastic and 644 non-sarcastic tweets, annotated by seven readers. We used the variant of the dataset employed by \cite{majumder2019sentiment}, which consists of 994 tweets (350 sarcastic and 644 non-sarcastic), excluding eye movement data.

\begin{figure}[h]
\center
\includegraphics[width=0.48\textwidth]{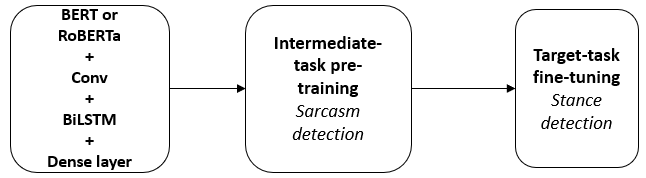}
\caption{The intermediate-task transfer learning pipeline.} 
\label{fig:exp_ppl}
\end{figure}

In this work, we implement two levels of transfer learning: first, from sarcasm detection to SD through intermediate-task pre-training; and second, from target-to-target through cross-target fine-tuning. The intermediate-task transfer learning pipeline is illustrated in Figure~\ref{fig:exp_ppl}.

\subsection{Underlying Model Architecture}

The model framework consists of an input layer, an embedding layer, and deep neural networks.

\subsubsection{Input Layer}
The input layer processes text \(x\) encoding stance information, comprising \(L\) words. The text \(x\) is converted into a vector of words and passed to the embedding layer.

\subsubsection{Embedding Layer}
We utilize BERT~\cite{devlin2018bert} and RoBERTa~\cite{liu2019roberta} for encoding textual input into hidden states \(H\). The choice of these language models is supported by their notable performance in the literature~\cite{savini2022intermediate}\cite{ghosh2019stance}\cite{phang2018sentence}\cite{pruksachatkun2020intermediate}\cite{ng2022my}\cite{gibthesis}.

\subsubsection{Deep Neural Networks}
\begin{figure*}[ht]
\center

\includegraphics[width=0.95\textwidth]{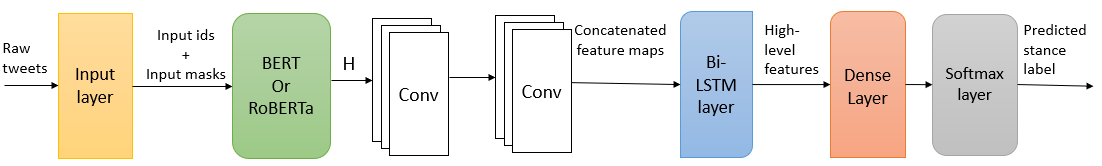}
\caption{The proposed model framework.} 
\label{fig:model}
\end{figure*}

The deep neural network module includes two convolutional layers (Conv), a BiLSTM layer, and a dense layer, which are applied on top of the embedding layer. The Conv layer identifies specific sequential word patterns within the text, creating a composite feature map from \(H\). This feature map aids the BiLSTM layer in capturing higher-level stance representations, which are further refined by the dense layer. The overall model framework is depicted in Figure~\ref{fig:model}.

\section{Experiments} \label{Exp}
This section delineates the datasets employed, details the data preprocessing procedures, outlines the baseline models, presents experimental results, and engages in a subsequent discussion.

\subsection{Datasets}
For evaluation purposes, we utilize two publicly available SD datasets: 1) the SemEval 2016 Task 6A Dataset (SemEval)~\cite{MohammadSK17}, and 2) the Multi-Perspective Consumer Health Query Data (MPCHI)~\cite{sen2018stance}.

\subsubsection{SemEval}
The SemEval dataset includes tweets manually annotated for stance towards specific targets, encompassing opinions and sentiments. For our experiments, we utilize tweets and their associated stance annotations. The dataset features tweets related to five distinct targets: Atheism (AT), Climate Change (CC), Feminist Movement (FM), Hillary Clinton (HC), and Legalization of Abortion (LA).

\subsubsection{MPCHI}
MPCHI is designed for stance classification to enhance Consumer Health Information (CHI) query search results. It comprises formal texts extracted from top-ranked articles corresponding to specific web search engine queries. The dataset includes sentences related to five distinct queries, which also serve as targets for stance classification: MMR vaccination and autism (MMR), E-cigarettes versus normal cigarettes (EC), Hormone Replacement Therapy post-menopause (HRT), Vitamin C and the common cold (VC), and sun exposure and skin cancer (SC).

\renewcommand{\arraystretch}{1.2} 

\begin{table*}[ht]
\caption{Original Statistics of the Datasets Divided into Training and Test Sets}
\begin{center}
\begin{small}
\begin{tabular}{|c||c||c|c|c||c|c|c|}
\hline
\multirow{2}{*}{\textbf{Dataset}} & \multirow{2}{*}{\textbf{Target}} & \multicolumn{3}{|c|}{\textbf{Training Samples}} & \multicolumn{3}{|c|}{\textbf{Test Samples}} \\
\cline{3-8}
& & INFAVOR & AGAINST & NONE & INFAVOR & AGAINST & NONE \\
\hline
\multirow{5}{*}{SemEval} & AT & 92 & 304 & 117 & 32 & 160 & 28 \\
\cline{2-8}
& CC & 212 & 15 & 168 & 123 & 11 & 35 \\
\cline{2-8}
& FM & 210 & 328 & 126 & 58 & 183 & 44 \\
\cline{2-8}
& HC & 112 & 361 & 166 & 45 & 172 & 78 \\
\cline{2-8}
& LA & 105 & 334 & 164 & 46 & 189 & 45 \\
\hline
\hline
\multirow{5}{*}{MPCHI} & MMR & 48 & 61 & 72 & 24 & 33 & 21 \\
\cline{2-8}
& SC & 68 & 51 & 117 & 35 & 26 & 42 \\
\cline{2-8}
& EC & 60 & 118 & 111 & 33 & 47 & 44 \\
\cline{2-8}
& VC & 74 & 52 & 68 & 37 & 16 & 31 \\
\cline{2-8}
& HRT & 33 & 95 & 44 & 9 & 41 & 24 \\
\hline
\end{tabular}
\end{small}
\end{center}
\label{tab:data}
\end{table*}

Each text in the datasets is annotated with one of three classes: \textit{InFavor}, \textit{Against}, and \textit{None}. Table~\ref{tab:data} presents the original statistical details of the datasets.

\subsection{Data Preprocessing}
We employ standard data preprocessing steps, including case folding, stemming, stop-word removal, and deletion of null entries across all datasets. Text normalization is performed following the method described by~\cite{han2011lexical}, and hashtag processing utilized Wordninja~\cite{wninja}. For neural network models relying on pre-trained embeddings, stemming and stop-word removal are omitted, as stemmed forms of terms may not be present in the pre-trained embeddings. The default tokenizer of the respective pre-trained language model is used to tokenize words in tweets prior to inputting them into the classifier.

\subsection{Baseline Models}
For the in-domain SD task, we evaluate our model against the top-performing results from the SemEval challenge~\cite{mohammad2016semeval}, as reproduced with minor modifications in~\cite{ghosh2019stance}. Additionally, we compare our model’s performance with recent SOTA methods in SD. The following first three baseline models are used for evaluating our model on the in-domain SD task, while the remaining models are used for evaluating the CTSD task.

\subsubsection{SemEval Models}
We select the Target-Specific Attention Neural Network (TAN-) proposed by~\cite{du2017stance} and the 1-D sem-CNN introduced by~\cite{kim2014convolutional} from the SemEval competition. Additionally, we include Com-BiLSTM and Com-BERT, implementations provided solely by~\cite{ghosh2019stance}.

\subsubsection{ChatGPT}
The work by~\cite{zhang2022would} explored the use of ChatGPT for SD by directly probing the generative language model to determine the stance of a given text, with a focus on specific targets from the SemEval task: FM, LA, and HC.

\subsubsection{Zero-Shot Stance Detection (ZSSD)}
The ZSSD technique~\cite{liang2022zero}, which employs contrastive learning, was implemented for the SemEval dataset similarly to ChatGPT.

\subsubsection{BiCond}
An LSTM model that uses bidirectional conditional encoding to learn both input text and target representations for SD~\cite{augenstein-etal-2016-stance}.

\subsubsection{TextCNN-E}
A variant of TextCNN~\cite{kim-2014-convolutional} adapted for the CTSD task by incorporating semantic and emotional knowledge into each word and expanding the dimensionality of each word vector~\cite{9746302}.

\subsubsection{Semantic-Emotion Knowledge Transferring (SEKT)}
This model leverages external semantic and emotion lexicons to facilitate knowledge transfer across different targets~\cite{zhang-etal-2020-enhancing-cross}.

\subsubsection{Target-Adaptive Pragmatics Dependency Graphs (TPDG)}
This model constructs two graphs: an in-target graph to capture inherent pragmatic dependencies of words for a specific target, and a cross-target graph to enhance the versatility of words across all targets~\cite{Liang2021TargetadaptiveGF}.

\subsubsection{Refined Meta-Learning (REFL)}
A SOTA CTSD model that utilizes meta-learning by refining the model with a balanced, easy-to-hard learning pattern and adapting it according to target similarities~\cite{9746302}.

\subsection{Experimental Settings}

The experimental setup adopts an inductive approach to transfer learning, where the target task model is initialized using parameters obtained from pre-training on sarcasm detection. This strategy is designed to enhance model performance for the target task. For the intermediate tasks, datasets are divided into training and validation sets solely for sarcasm detection pre-training. Given that Sav2C and ST are the smallest intermediate-task datasets, five-fold cross-validation is utilized for these, while SARC, being larger, employs an 80/20 train/validation split. In contrast, the target task featured a separate test set for final evaluations and comparisons.

Consistent with the methodologies of~\cite{ghosh2019stance}, datasets are divided into training and test sets using similar proportions for in-domain  SD, while CTSD employs a leave-one-out strategy. In this approach, data from all source targets are used for model training, and the test data for the destination target is reserved for model evaluation. Each SD dataset consists of five targets; thus, during CTSD experimentation, four targets are used for training, and the remaining target is used for evaluation. Table~\ref{tab:data_ctsd} details the statistics of the datasets after incorporating the experimental settings of CTSD.

\renewcommand{\arraystretch}{1.2} 

\begin{table*}[ht]
\caption{ Statistics of the datasets after Incorporating Cross-Target Stance Detection}
\begin{center}
\begin{small}
\begin{tabular}{|c||c||c|c|c||c|c|c|}
\hline
\multirow{2}{*}{\textbf{Dataset}}& \multirow{2}{*}{\textbf{Target}}&\multicolumn{3}{|c|}{\textbf{Training samples}} &\multicolumn{3}{|c|}{\textbf{Test samples}}\\
\cline{3-8} 
\textbf{}& &INFAVOR & AGAINST&NONE  &INFAVOR &AGAINST &NONE  \\
\hline
\multirow{5}{*}{SemEval} &AT &910 &1593 &826 &32 &160 &28 \\
\cline{2-8}
&CC &699 &2031 &767 &123 &11&35\\
\cline{2-8}
&FM  &766 &1546 &800 &58 &183 &44\\
\cline{2-8}
&HC &878 &1524 &726 &45 &172 &78\\
\cline{2-8}
&LA &883 &1534 &761 &46 &189 &45\\
\hline
\hline
\multirow{5}{*}{MPCHI} &MMR &314 &402 &425 &24 &33 &21\\
\cline{2-8}
&SC &279 &417 &365 &35 &26 &42\\
\cline{2-8}
&EC &301 &343 &376 &33 &47 &44\\
\cline{2-8}
&VC &276 &424 &421 &37 &16 &31\\
\cline{2-8}
&HRT &342 &358 &453 &9 &41 &24\\
\hline
\end{tabular}
\end{small}
\end{center}
\label{tab:data_ctsd}
\end{table*}

The Conv layer uses a kernel size of 3 with 16 filters and a ReLU activation function. A BiLSTM layer with a hidden state of 768, corresponding to the hidden state size of the pre-trained language models, is employed. The dense layer has an output size of 3 and utilizes a softmax activation function. All experiments are conducted on an NVIDIA Quadro RTX 4000 GPU.

Hyperparameter tuning involves multiple experiments to select the optimal intermediate-task training scheme based on results from a holdout development set. The best-performing per-task model is then evaluated on the test set. The training process uses a mini-batch size of 16 samples and the Adam optimizer~\cite{kingma2014adam}, with cross-entropy loss as the cost function. Training epochs ranges from 10 to 50, with early stopping applied if validation accuracy on holdout data plateaus for five consecutive epochs. The learning rate is initially set to 3e-5, decaying to 1e-9 for the intermediate task and 1e-10 for the target task. A dropout rate of 0.25 is introduced between model layers to mitigate overfitting. To address class imbalance, class weights are incorporated during training to improve generalization for underrepresented classes. Experimental setups adhere to the configurations outlined in the original papers for baseline models unless otherwise specified, in which case our experimental configurations are applied.

\subsection{Evaluation Metrics}

In alignment with previous studies~\cite{ghosh2019stance}\cite{augenstein2016stance}\cite{mohammad2016semeval}, the evaluation of our model is based on the average macro F1-score for the \textit{InFavor} and \textit{Against} classes.

\subsection{Results}

We first present the results for in-domain SD, followed by the CTSD results. Baseline results for CTSD are referenced from~\cite{9746302}. All results are averaged over five experimental runs per target task.

Table~\ref{tab:res1} displays the experimental outcomes for in-domain SD before the incorporation of sarcasm detection pre-training. Results for ChatGPT and ZSSD are directly transcribed from their original publications, while other baseline results are replicated in our experiments. The table demonstrates the notable performance of our BERT-based model across various targets, achieving superior results in most metrics except HC and CC, where ChatGPT and our RoBERTa-based model excel. Consequently, we select our BERT-based model for subsequent experiments.

\renewcommand{\arraystretch}{1.2} 

\begin{table*}[ht]
\caption{Experimental results without sarcasm detection pre-training}
\begin{center}
\begin{small}
\begin{tabular}{|l||c|c|c|c|c|c||c|c|c|c|c|c|}
\hline
\multirow{2}{*}{\textbf{Model}}&\multicolumn{6}{|c|}{SemEval}&\multicolumn{6}{|c|}{MPCHI}\\
\cline{2-13} 
&\textbf{AT}&\textbf{CC}&\textbf{FM}&\textbf{HC}&\textbf{LA}&\textbf{Avg}&\textbf{MMR}&\textbf{SC}&\textbf{EC}&\textbf{VC}&\textbf{HRT}&\textbf{Avg}\\
\hline
Sem-TAN-&0.596 &0.420 &0.495 &0.543 &0.603 &0.531 &0.487 &0.505 &0.564 &0.487 &0.467 &0.502\\

Sem-CNN&0.641 &0.445 &0.552 &0.625 &0.604 &0.573 &0.524 &0.252 &0.539 &0.524 &0.539 &0.476 \\ 

Com-BiLSTM&0.567 &0.423 &0.508 &0.533 &0.546 &0.515 &0.527 &0.522 &0.471 &0.474 &0.469 &0.493 \\ 

ZSSD &0.565 &0.389 &0.546 &0.545 &0.509 &0.511 &- &- &- &- &- &-\\

Com-BERT&0.704 &0.466 &0.627 &0.620 &0.673 &0.618 &0.701 &0.691 &0.710 &0.617 &0.621 &0.668\\

ChatGPT &- &- &0.690 &\textbf{0.780} &0.593 &0.687 &- &- &- &- &- &-\\

Ours-RoBERTa&0.740 &\textbf{0.775} &0.689 &0.683 &0.696 &0.712 &0.692 &0.687 &0.700 &0.701 &0.698&0.695 \\

Ours-BERT&\textbf{0.767} &0.755 &\textbf{0.697} &0.704 &\textbf{0.702} &\textbf{0.725} &\textbf{0.747} &\textbf{0.722 } &\textbf{0.704} &\textbf{0.702 } &\textbf{0.732} &\textbf{0.721} \\ 
\hline
\end{tabular}
\end{small}
\end{center}
\label{tab:res1}
\end{table*}

\renewcommand{\arraystretch}{1.2} 

\begin{table*}[ht]
\caption{Experimental results with sarcasm-detection pre-training}
\begin{center}
\begin{small}
\begin{tabular}{|l||c|c|c|c|c|c||c|c|c|c|c|c|}
\hline
\multirow{2}{*}{\textbf{Task}}&\multicolumn{6}{|c|}{SemEval}&\multicolumn{6}{|c|}{MPCHI}\\
\cline{2-13} 
&\textbf{AT} &\textbf{CC} &\textbf{FM} &\textbf{HC} &\textbf{LA} &\textbf{Avg} &\textbf{MMR} &\textbf{SC} &\textbf{EC} &\textbf{VC} &\textbf{HRT} &\textbf{Avg}\\
\hline
SaV2C &0.595  &0.718  &0.596  &0.645  &0.578  &0.626 &0.605 &0.545 &0.545 &0.352 &0.495 &0.508 \\

SARC &0.697  &0.612  &0.683  &0.557  &0.641  &0.638 &0.605
&0.545 &0.545 &0.352 &0.495 &0.508\\ 

ST & \textbf{0.769} &\textbf{0.800 }  & \textbf{0.774} &\textbf{0.795}  &\textbf{0.741}  &\textbf{0.775} &\textbf{0.749} &\textbf{0.727} &\textbf{0.704} &\textbf{0.703} &\textbf{0.739} &\textbf{0.724}\\
\hline
\end{tabular}
\end{small}
\end{center}
\label{tab:res2}
\end{table*}

\renewcommand{\arraystretch}{1.2} 

\begin{table*}[ht]
\caption{Experimental results of Cross-Target Stance Detetcion with sarcasm-detection pre-training}
\begin{center}
\begin{small}
\begin{tabular}{|l||c|c|c|c|c|c||c|c|c|c|c|c|}
\hline
\multirow{2}{*}{\textbf{Task}}&\multicolumn{6}{|c|}{SemEval}&\multicolumn{6}{|c|}{MPCHI}\\
\cline{2-13} 
&\textbf{AT} &\textbf{CC} &\textbf{FM} &\textbf{HC} &\textbf{LA} &\textbf{Avg} &\textbf{MMR} &\textbf{SC} &\textbf{EC} &\textbf{VC} &\textbf{HRT} &\textbf{Avg}\\
\hline
BiCond &0.526 &0.512 &0.527 &0.536 &0.493  &0.519 &- &- &- &- &- &- \\
TextCNN-E &0.534 &0.633 &0.582 &0.591 &0.550 &0.578 &- &- &- &- &- &-\\ 

SEKT &0.623 &0.600 &0.648 &- &0.649  &0.630 &- &- &- &- &- &-\\ 
TPDG &0.654 &0.667 &0.669 &0.630 &0.600  &0.644 &- &- &- &- &- &-\\ 
REFL &0.650 &0.671 &\textbf{0.734} &0.652 &0.623  &0.666 &- &- &- &- &- &-\\ 
Ours & \textbf{0.689} &\textbf{0.697 }  & 0.730 &\textbf{0.682}  &\textbf{0.656}  &\textbf{0.691} &\textbf{0.699} &\textbf{0.687} &\textbf{0.695} &\textbf{0.701} &\textbf{0.700} &\textbf{0.696}\\
\hline
\end{tabular}
\end{small}
\end{center}
\label{tab:res_ctsd}
\end{table*}

\renewcommand{\arraystretch}{1.2} 

\begin{table*}[h]
\caption{Experimental results of an ablation study}
\begin{center}
\begin{small}
\begin{tabular}{|l||c|c|c|c|c|c||c|c|c|c|c|c|}
\hline
\multirow{2}{*}{\textbf{Model}}&\multicolumn{6}{|c|}{SemEval}&\multicolumn{6}{|c|}{MPCHI}\\
\cline{2-13} 
&\textbf{AT} &\textbf{CC} &\textbf{FM} &\textbf{HC} &\textbf{LA} &\textbf{Avg} &\textbf{MMR} &\textbf{SC} &\textbf{EC} &\textbf{VC} &\textbf{HRT} &\textbf{Avg}\\
\hline
BERT &0.674  &0.677 &0.678 &0.609 &0.685 &0.665 &0.568 &0.519 &0.441 &0.482 &0.595 &0.521\\

BERT+Conv+BiLSTM &0.767 &0.755 &0.697 &0.704 &0.702 &0.725 &0.747 &0.722 &\textbf{0.704} &0.702 &0.732 &0.721\\ 

ST+BERT &0.712 &0.735 &0.698 &0.687 &0.696 &0.706 &0.687 &0.601 &0.540 &0.466 &0.546 &0.568\\

ST+BERT+Conv&\textbf{0.770} &0.759 &0.689 &0.683 &0.694 &0.719 &0.458 &0.535 &0.479 &0.350 &0.524 &0.469\\

ST+BERT+BiLSTM &0.747 &0.765 &0.675 &0.657 &0.678 &0.704 &0.640
&0.618 &0.573 &0.528 &0.633 &0.598\\

ST+BERT+Conv+BiLSTM & 0.769 &\textbf{0.800 }  & \textbf{0.774} &\textbf{0.795}  &\textbf{0.741}  &\textbf{0.775} &\textbf{0.749} &\textbf{0.727} &\textbf{0.704} &\textbf{0.703} &\textbf{0.739} &\textbf{0.724}\\
\hline
\end{tabular}
\end{small}
\end{center}
\label{tab:res3}
\end{table*}

Table~\ref{tab:res2} reports the results of incorporating sarcasm detection pre-training with our model for in-domain SD. Performance improves by \textbf{0.550} on SemEval and \textbf{0.003} on MPCHI when pre-training with ST, surpassing all baseline models listed in Table~\ref{tab:res1}. However, performance decreases with Sav2C and SARC. Therefore, subsequent results utilize the ST model.

Table~\ref{tab:res_ctsd} presents the results of CTSD. Notably, no baseline models have been evaluated on the MPCHI dataset, focusing instead on SemEval with one target not addressed by the SEKT baseline. Our model outperforms all CTSD models listed in the table on the average macro F1 measure.

Table~\ref{tab:res3} summarizes the results of an ablation study on the in-domain task. Various base model components are systematically excluded to evaluate their contributions to the overall model framework. The model integrating all components—BERT, Conv, BiLSTM, and sarcasm pre-training—achieves the highest average F1-scores of \textbf{0.775} and \textbf{0.724} on SemEval and MPCHI, respectively.

\subsection{Failure Analysis and Discussion} \label{Dis}

Following the results presented in Table~\ref{tab:res1}, a detailed failure analysis is conducted to investigate the misclassified test samples. The analysis reveals that misclassifications in the SemEval dataset are predominantly associated with texts containing sarcastic content, consistent with prior findings~\cite{ghosh2019stance}. This observation supports the rationale for incorporating sarcasm-detection pre-training prior to fine-tuning for SD. Conversely, misclassifications in the MPCHI dataset are primarily linked to samples that contained large, generic health-related facts that are neutral with respect to the target under study. Additional insights derived from the experiments and results across all tasks are discussed below.

\subsubsection{Performance of Our Model Relative to SOTA Models Without Sarcasm Detection}

Our model demonstrates superior performance compared to SOTA models even in the absence of sarcasm detection. Specifically, it outperforms ChatGPT and Com-BERT, which are among the top-performing models, on both SemEval and MPCHI by \textbf{0.038} and \textbf{0.053} in average F1-scores, respectively, for the in-domain SD task. While Com-BERT utilizes only BERT and a dense layer for classification, our model benefits from additional Conv and BiLSTM layers preceding the dense layer, which contributes to the observed performance improvement. Furthermore, the inclusion of the BiLSTM module in our model results in better performance compared to using pooling layers after the Conv module. This finding highlights the effectiveness of our model architecture in capturing nuanced representations, leading to improved generalization for SD tasks.

\subsubsection{Correlation Between Sarcasm Detection and SD}

An illustrative example of misclassification involves the statement: ``\textit{I like girls. They just need to know their place. \#SemST}", a sarcastic comment from the FM target in SemEval. The true label for this example is \textit{Against}, but it is misclassified as \textit{InFavor} before the incorporation of sarcasm-detection pre-training. Notably, sarcastic samples in the \textit{Against} class are often misclassified as \textit{InFavor} due to their overtly positive content. After integrating sarcasm detection through pre-training, 85\% of these misclassified sarcastic samples are correctly predicted. This result underscores the importance of sarcasm-detection pre-training in enhancing the performance of SD models.

\subsubsection{Challenges in Using Sarcasm Detection Models for Intermediate-Task Transfer Learning on SD}

The integration of SARC and SaV2C knowledge into the model pipeline introduces noise and adversely affects model performance on SD compared to using ST knowledge. Analysis of Sav2C and SARC reveals several discrepancies with the target task. For instance, the average sentence length in Sav2C and SARC is longer compared to SemEval and MPCHI samples. Additionally, SARC is sourced from different domains than SemEval and MPCHI, leading to variations in topic coverage, vocabulary overlap, and the framing of ideas. SARC, being the largest intermediate task, spans a wide range of topics across various subreddits, while ST, which performs best, shares a similar average sentence length with the target tasks and is also crowd-sourced from Twitter (X). This alignment likely contributes to the superior performance observed when using ST as an intermediate task for SemEval. Consequently, the mismatched attributes of certain intermediate tasks can negatively impact model performance. This underscores the need for careful selection and experimentation when choosing a sarcasm model for transfer learning in SD.

\subsubsection{Performance of Cross-Target Stance Detection}

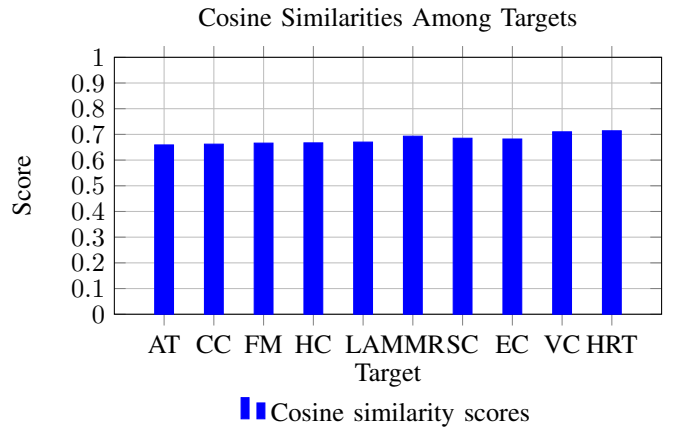
\begin{figure}[ht] 
    \centering
    \begin{tikzpicture}
        \begin{axis}[
            title={Cosine Similarities Among Targets},
            xlabel={Target},
            ylabel={Score},
            xmin=0, xmax=11,
            ymin=0, ymax=1,
            xtick={1,2,...,10}, 
            xticklabels={AT, CC, FM, HC, LA, MMR, SC, EC, VC, HRT}, 
            ytick={0,0.1,0.2,0.3,0.4,0.5,0.6,0.7,0.8,0.9,1.0}, 
            legend style={at={(0.5,-0.3)}, anchor=north, draw=none}, 
            grid=major,
            width=\columnwidth, 
            height=5cm, 
            ybar, 
            bar width=0.25cm, 
        ]
        \addplot[color=blue, fill=blue] coordinates {
            (1,0.659) (2,0.662) (3,0.666) (4,0.667) (5,0.670) (6,0.693) (7,0.685) (8,0.682) (9,0.710) (10,0.714)
        };
        \legend{Cosine similarity scores}
        \end{axis}
    \end{tikzpicture}
    \caption{Cosine similarity scores for each target in comparison with other targets within their respective datasets.}
    \label{fig:bargraph}
\end{figure}

The CTSD task exhibits comparable performance to the in-domain task, despite using out-of-domain data during model fine-tuning. This suggests that our model effectively learns common features from various targets, thereby leveraging this data to perform well on new targets in CTSD. To further understand this observation, cosine similarity scores on the pre-trained BERT embeddings are analyzed. Figure~\ref{fig:bargraph} illustrates the cosine similarities between each target and the other targets in their respective datasets. In the figure, LAMMRSC should read as LA, MMR, and SC on the X axis. The figure demonstrates that all targets share common vocabulary with others, leading to shared features. Additionally, MPCHI targets have higher cosine similarity scores than SemEval targets, which aligns with the superior CTSD performance observed on the MPCHI task.

\subsubsection{Ablation Study on Sarcasm Knowledge}

The results of the ablation study presented in Table~\ref{tab:res3} provide insights into the contribution of each module and the overall impact of sarcasm detection pre-training on SD performance. Comparing the results in Table~\ref{tab:res1} and Table~\ref{tab:res3}, the incorporation of sarcasm knowledge significantly enhances model performance on the SemEval task compared to the MPCHI task. SemEval includes a large volume of opinionated and sarcastic texts, whereas the MPCHI dataset primarily consists of health-related facts, with occasional sarcastic expressions. As a result, there is a modest improvement in performance on MPCHI when sarcasm detection is used. This suggests the potential for exploring BERT or RoBERTa embeddings pre-trained on health-related data specifically for SD on MPCHI as a promising avenue for future research.

\section{Limitations}\label{limit}

Despite the significant contributions of this study to NLP in social media contexts, several limitations warrant consideration. Firstly, the extent of model performance improvement is dependent on the characteristics of both the intermediate sarcasm detection task and the ultimate SD task. Variations in linguistic features across datasets used for sarcasm detection and SD may limit the generalizability of the study’s findings. Secondly, while the integration of BERT or RoBERTa with other deep-learning methodologies represents an innovative approach, the complexity of the model architecture may pose challenges in terms of computational resources and interoperability in certain contexts. Thirdly, the CTSD task presents additional challenges, as the language models employed may not be compatible across different targets. Lastly, the heavy reliance on fine-tuning techniques and specific datasets raises concerns about the model's ability to generalize effectively across diverse text types or domains not covered within the training data.

\section{Conclusion and Future Work} \label{Con}

In this study, we have proposed a transfer-learning framework that integrates sarcasm detection for SD. We have utilized pre-trained language models, RoBERTa and BERT, which have been individually fine-tuned and subsequently concatenated with other deep neural networks, with BERT demonstrating particularly promising results. The model has been pre-trained on three sarcasm-detection tasks before being fine-tuned on two target SD tasks. Our evaluations, including in-domain SD and CTSD, have shown that our approach outperformed SOTA models, even before incorporating sarcasm knowledge. The correlation between sarcasm detection and SD has been established, with the integration of sarcasm knowledge significantly enhancing model performance; notably, 85\% of misclassified samples in the SemEval task have been accurately predicted after incorporating sarcasm knowledge. Failure analysis has indicated that the SemEval dataset, rich in opinionated sarcastic samples, has benefited significantly from sarcasm pre-training, in contrast to the MPCHI dataset, which primarily consists of generic health-related facts. Furthermore, our study has revealed that not all intermediate sarcasm-detection tasks have improved SD performance due to mismatched linguistic attributes. Additionally, the CTSD task has demonstrated performance on par with the in-domain task despite using a zero-shot fine-tuning approach, effectively addressing the issue of limited annotated samples from new targets. Finally, the ablation study has highlighted that the optimal performance of the model is achieved when all components are utilized.

To the best of our knowledge, this work represents the inaugural application of sarcasm-detection pre-training within a BERT (RoBERTa)+Conv+BiLSTM architecture before fine-tuning for SD. Our approach serves as a foundational reference, setting a baseline for future research in this domain. Future work will explore variant BERT or RoBERTa embeddings tailored to health-related text data for the MPCHI task and will focus on a more comprehensive evaluation of other intermediate tasks, including sentiment and emotion knowledge.

\section*{ACKNOWLEDGMENT}
This work is supported by the National Science Foundation (NSF) under Award number OIA-1946391, Data Analytics that are Robust and Trusted (DART). We sincerely thank our  anonymous reviewers for their valuable insights and constructive feedback. Additionally, we extend our gratitude to all individuals who contributed to this study in various capacities.



\end{document}